# Impact of Deep Learning Libraries on Online Adaptive Lightweight Time Series Anomaly Detection

Ming-Chang Lee[1] and Jia-Chun Lin[2]

[1]Department of Computer science, Electrical engineering and Mathematical sciences, Høgskulen på Vestlandet (HVL), Bergen, Norway

[2]Department of Information Security and Communication Technology, Norwegian University of Science and Technology, Gjøvik, Norway

[1] mingchang1109@gmail.com
[2] jia-chun.lin@ntnu.no

# Impact of Deep Learning Libraries on Online Adaptive Lightweight Time Series Anomaly Detection

Ming-Chang Lee[1] [a] and Jia-Chun Lin[2] [b]

[1]*Department of Computer science, Electrical engineering and Mathematical sciences, Høgskulen på Vestlandet (HVL), Bergen, Norway*
[2]*Department of Information Security and Communication Technology, Norwegian University of Science and Technology (NTNU), Gjøvik, Norway*
mingchang1109@gmail.com, jia-chun.lin@ntnu.no

Keywords: Time series, univariate time series, anomaly detection, online model training, unsupervised learning, TensorFlow, Keras, PyTorch, Deeplearning4j

Abstract: Providing online adaptive lightweight time series anomaly detection without human intervention and domain knowledge is highly valuable. Several such anomaly detection approaches have been introduced in the past years, but all of them were only implemented in one deep learning library. With the development of deep learning libraries, it is unclear how different deep learning libraries impact these anomaly detection approaches since there is no such evaluation available. Randomly choosing a deep learning library to implement an anomaly detection approach might not be able to show the true performance of the approach. It might also mislead users in believing one approach is better than another. Therefore, in this paper, we investigate the impact of deep learning libraries on online adaptive lightweight time series anomaly detection by implementing two state-of-the-art anomaly detection approaches in three well-known deep learning libraries and evaluating how these two approaches are individually affected by the three deep learning libraries. A series of experiments based on four real-world open-source time series datasets were conducted. The results provide a good reference to select an appropriate deep learning library for online adaptive lightweight anomaly detection.

## 1 INTRODUCTION

A time series refers to a sequence of data points indexed in time order, and it is a collection of observations obtained via repeated measurements over time (Ahmed et al., 2016). Examples of time series include stock prices, retail sales, electricity consumption, temperatures, humidity, CO2, blood pressures, heart rates, etc. Due to the increasing prevalence of the Internet of Things (IoT), more and more different time series are continuously generated by diverse IoT sensors and devices over time. Analyzing time series is valuable to businesses and organizations since it gives insight into what has happened and identifies trends and seasonal variances to aid in the forecasting of future events. It also enables businesses and organizations to take appropriate policies or make better decisions (Kieu et al., 2018; Yatish and Swamy, 2020).

Time series anomaly detection is an analysis task focusing on detecting anomalous or abnormal data points in time series, and it has been widely used in various applications ranging from cloud systems (Deka et al., 2022), smart grids (Zhang et al., 2021), healthcare (Pereira and Silveira, 2019) to agriculture (Moso et al., 2021). Many time series anomaly detection approaches have been introduced in the last decade. Some were designed for univariate time series where there is only one time-dependent variable, and the other approaches were designed for multivariate time series that consists of more than one time-dependent variables. In this paper, we focus on the studies for univariate time series. To be more specific, we focus on univariate time series anomaly detection approaches that possess the following features: Unsupervised learning, online model training, adaptability, and lightweight since these features decide whether an approach is practical or not. (Blázquez-García et al., 2021).

Unsupervised learning refers to machine learning models that have a self-learning ability to draw inference from a dataset containing a small minority of abnormal data without any label. Since most of real-world time series data do not have any label, it is desirable to have an unsupervised anomaly detection approach. Conventional machine learning models are usually trained with a pre-collected dataset in an offline manner. Once the models are trained, they are used for inference without any change. Hence, they cannot reflect unseen situations or adapt to changes on time series (Eom et al., 2015). Unlike offline model training, online model training enables a machine learning model to be trained on the fly, implying that the model can adapt to changes in the pattern of the time series (i.e., adaptability). This feature is getting more and more popular, and it has been provided by some systems or approaches such as (Lee et al., 2020b; Eom et al., 2015; Chi et al., 2021). Finally, lightweight means that an anomaly detection approach neither has a complex network structure/design nor requires excessive computation resources such as General-Purpose Graphics processing units (GPGPUs) or high-performance computers.

---

[a] https://orcid.org/0000-0003-2484-4366
[b] https://orcid.org/0000-0003-3374-8536

According to our survey, only few state-of-the-art approaches satisfy all the above-mentioned characteristics, such as RePAD (Lee et al., 2020b), ReRe (Lee et al., 2020a), SALAD (Lee et al., 2021b), and RePAD2 (Lee and Lin, 2023). However, all of them were only implemented in one specific deep learning library. In fact, a number of deep learning (DL) libraries have been introduced and widely used, such as TensorFlow (Abadi et al., 2016), PyTorch (Paszke et al., 2019), and Deeplearning4j (Deeplearning4j, 2023). They have a common goal to facilitate the complicated data analysis process and offer integrated environments on top of standard programming languages (Nguyen et al., 2019). However, it is unclear the impact of these DL libraries on online adaptive lightweight anomaly detection.

Therefore, this paper focuses on investigating how different DL libraries affect online adaptive lightweight time series anomaly detection by implementing two state-of-the-art anomaly detection approaches in three widely-used deep learning libraries. It is worth noting that our focus is not to compare different time series anomaly detection approaches regarding their detection accuracy or response time. Instead, we emphasize on investigating how these approaches are individually affected by different DL libraries.

A series of experiments based on open-source time series datasets were performed. The results show that DL libraries have a great impact on not only anomaly detection accuracy but also response time. Therefore, it is important to take the selection of DL libraries into consideration when one would like to design and implement an online adaptive lightweight time series anomaly detection approach.

The rest of the paper is organized as follows: Section 2 describes time series anomaly detection approaches and DL libraries. Section 3 gives an overview of the related work. Section 4 introduces evaluation setup. Section 5 presents the evaluation results. Section 6 concludes this paper and outlines future work.

## 2 BACKGROUND

In this section, we introduce state-of-the-art anomaly detection approaches for univariate time series and some well-known DL libraries.

### 2.1 Anomaly Detection Approaches for Univariate Time Series

Existing anomaly detection approaches for univariate time series can be roughly classified into two categories: statistical based and machine learning based. Statistical-based anomaly detection approaches attempt to create a statistical model for normal time series data and use this model to determine if a data point is anomalous or not. Example approaches include AnomalyDetectionTs and AnomalyDetectionVec proposed by Twitter (Twitter, 2015), and Luminol introduced by LinkedIn (LinkedIn, 2018). However, statistical-based approaches might not perform well if the data does not follow a known distribution (Alimohammadi and Chen, 2022).

On the other hand, machine learning based approaches attempt to detect anomalies without assuming a specific generative model based on the fact that it is unnecessary to know the underlying process of the data (Braei and Wagner, 2020). Greenhouse (Lee et al., 2018) is a time series anomaly detection algorithm based on Long Short-Term Memory (LSTM), which is a special recurrent neural network suitable for long-term dependent tasks (Hochreiter and Schmidhuber, 1997). Greenhouse adopts a Look-Back and Predict-Forward strategy to learn the distribution of the training data. For a given time point, a window of most recently observed data point values are used to predict future data point values. However, Greenhouse is not an online approach since its LSTM model is trained with a pre-collected training data. Besides, it requires users to determine a proper detection threshold.

RePAD (Lee et al., 2020b) is an online real-time lightweight unsupervised time series anomaly detection approaches based on LSTM and the Look-Back and Predict-Forward strategy. RePAD utilizes a simple LSTM network (with only one hidden layer and ten hidden units) to train a LSTM model with short-term historical data points, predict each upcoming data point, and then decide if each data point is anomalous based on a dynamically calculated detection threshold. Different from Greenhouse, RePAD does not need to go through any offline training. Instead, RePAD trains its LSTM model on the fly. RePAD will keep using the same LSTM model if the model predicts well. When the prediction error of the model is higher than or equal to a dynamically calculated detection threshold, RePAD will retrain another new model with recent data points.

ReRe (Lee et al., 2020a) is an enhanced time series anomaly detection based on RePAD, and it was designed to further reduce false positive rates. ReRe utilizes two LSTM models to jointly detect anomalous data points. One model works exactly like RePAD, whereas the other model works similar to RePAD but with a stricter detection threshold. Compared with RePAD, ReRe requires more compute resources due to the use of two LSTM models.

SALAD (Lee et al., 2021b) is another online self-adaptive unsupervised time series anomaly detection approach designed for time series with a recurrent data pattern, and it is also based on RePAD. Different from RePAD, SALAD consists of two phases. The first phase converts the target time series into a series of average absolute relative error (AARE) values on the fly. The second phase predicts an AARE value for every upcoming data point based on short-term historical AARE values. If the difference between a calculated AARE value and the corresponding forecast AARE value is higher than a self-adaptive detection threshold, the corresponding data point is considered anomalous.

Ziu et al. (Niu et al., 2020) introduced LSTM-based VAE-GAN, which stands for a Long Short-Term Memory-based variational autoencoder generation adversarial networks. This method consists of one offline training stage to learn the distribution of normal time series, and one anomaly detection stage to calculate anomaly score for each data point in the target time series. This method jointly trains the encoder, the generator, and the discriminator to take advantage of the mapping ability of the encoder and the discriminatory ability of the discriminator. However, the method requires that the training data contains no anomalies. Besides, the method is not an online approach since its detection model will not be retrained or updated after the training stage, meaning that it is not adaptive.

Ibrahim et al. (Ibrahim et al., 2022) proposed a hybrid deep learning approach that combines one-dimensional convolutional neural network with bidirectional long short-term memory (BiLSTM) for anomaly detection in univariate time series. However, the approach requires offline training and considerable training time due to parameter tuning required by the used hybrid approach.

## 2.2 Deep Learning Libraries

Over the last few years, machine learning has seen significant advances. Many different machine learning algorithms have been introduced to address different problems. In the meantime, many DL libraries have been developed by academy, industry, and open-source communities, attempting to provide a fair abstraction on the ground complex tasks with simple functions that can be used as tools for solving larger problems (Ketkar and Santana, 2017).

TensorFlow (Abadi et al., 2016) is a popular open-source Python-based DL library created and maintained by Google. It uses dataflow graphs to represent both the computation in an algorithm and the state on which the algorithm operates. TensorFlow is designed for large-scale distributed training and inference. It can run on a single CPU system, GPUs, mobile devices, and large-scale distributed systems. However, its low-level application programming interface (API) makes it difficult to use (Nguyen et al., 2019). Because of this, TensorFlow is usually used in combination with Keras (Keras, 2023), which is a Python wrapper library providing high-level, highly modular, and user-friendly API.

CNTK (CNTK, 2023) stands for Cognitive Toolkit, and it was introduced by Microsoft and written in C++ programming language. It supports the Open Neural Network Exchange (ONNX) format, allowing easy model transformation from one DL library to another one. As compared with TensorFlow, CNTK is less popular (Nguyen et al., 2019). Moreover, the official website of CNTK shows that CNTK is no longer actively developed.

PyTorch (Paszke et al., 2019) is an open-source DL framework based on the Torch library. It aims to provide an easy to use, extend, develop, and debug framework. It is equipped with a high-performance C++ runtime that developers can leverage for production environments while avoiding inference via Python (Ketkar and Santana, 2017). PyTorch supports tensor computation with strong GPU acceleration and allows a network to change the way it behaves with small effort using dynamic computational graphs. Similar to CNTK, it also supports the ONNX format.

Deeplearning4j is an open source distributed deep learning library released by a startup company called Skymind in 2014 (Deeplearning4j, 2023)(Wang et al., 2019). Deeplearning4j is written for java programming language and java virtual machine (JVM). It is powered by its own open-source numerical computing library called ND4J, and it supports both CPUs and GPUs. Deeplearning4j provides implementations of the restricted Boltzmann machine, deep belief net, deep autoencoder, recurrent neural network, word2vec, doc2vec, etc.

## 3 RELATED WORK

Nguyen et al. (Nguyen et al., 2019) conducted a survey on several DL libraries. They also analyzed strong points and weak points for each library. However, they did not conduct any experiments to compare these DL libraries. Wang et al. (Wang et al., 2019) compared several DL libraries in terms of model design ability, interface property, deployment ability, performance, framework design, and development prospects by using some benchmarks. The authors also made suggestions about how to choose DL frameworks in different scenarios. Nevertheless, their general evaluation and analysis are unable to answer the specific question that this paper attempts to answer, i.e., how DL libraries affect online adaptive lightweight time series anomaly detection approaches.

Kovalev et al. (Kovalev et al., 2016) evaluated the training time, prediction time, and classification accuracy of a fully connected neural network (FCNN) under five different DL libraries: Theano with Keras, Torch, Caffe, Tensorflow, and Deeplearning4j. Apparently, their results are not applicable to lightweight anomaly detection approaches.

Zhang et al. (Zhang et al., 2018) evaluated the performance of several state-of-the-art DL libraries, including TensorFlow, Caffe2, MXNet, PyTorch and TensorFlow Lite on different kinds of hardware, including MacBook, FogNode, Jetson TX2, Raspberry Pi, and Nexus 6P. The authors chose a large-scale convolutional neural network (CNN) model called AlexNet (Krizhevsky et al., 2017) and a small-scale CNN model called SqueezeNet (Iandola et al., 2016), and evaluated how each of them performs under different combination of hardware and DL libraries in terms of latency, memory footprint, and energy consumption. According to the evaluation results, there is no single winner on every metric since each has its own metric. Due to the fact that two used CNN models are much complex than lightweight anomaly detection approaches, their evaluation results and suggestions may not be applicable.

Zahidi et al. (Zahidi et al., 2021) conducted an analysis to compare different Python-based and Java-based DL libraries and to see how they support different natural language processing (NLP) tasks. Due to the difference between NLP tasks and time series analysis, their results still cannot be applied to the work of this paper.

Zhang et al. (Zhang et al., 2022) built a benchmark that includes six representative DL libraries on mobile devices (TFLite, PyTorchMobile, ncnn, MNN, Mace, and SNPE) and 15 DL models (10 of them are for image classification, 3 of them are for object detection, 1 for semantic segmentation, and 1 for text classification). The authors then performed a series of experiments to evaluate the performance of these DL libraries on the 15 DL models and different mobile devices. According to their analysis and observation, there is no DL libraries that perform best on all tested scenarios and that the impacts of DL libraries may overwhelm DL algorithm design and hardware capacity. Apparently, the target of our paper is completely different from that of Zhang et al.'s paper. Even though their results point out some useful conclusions, their results cannot help us get a clear answer about how different DL libraries affect online adaptive lightweight anomaly detection.

## 4 EVALUATION SETUP

Based on the description in the Background section, we chose RePAD and SALAD to be our target anomaly detection approaches because both of them possess all previously mentioned desirable features (i.e., unsupervised learning, online model training, adaptability, and lightweight). As for DL libraries, we chose TensorFlow-Keras, PyTorch, and Deeplearning4j because they are popular and widely used. Recall both TensorFlow-Keras and PyTorch are based on Python, it would be interesting to see how Deeplearning4j performs as compared with TensorFlow-Keras and PyTorch. Here, the versions of TensorFlow-Keras, PyTorch, and Deeplearning4j are 2.9.1, 1.13.1, and 0.7-SNAPSHOT, respectively.

We implemented RePAD and SALAD in the three DL libraries. Hence, there are six combinations as shown in Table 1. RePAD-TFK refers to RePAD implemented in TensorFlow-Keras, SALAD-PT refers to SALAD implemented in PyTorch, and so on so forth.

Table 1: The six combinations studied in this paper.

|  | RePAD | SALAD |
|---|---|---|
| TensorFlow-Keras | RePAD-TFK | SALAD-TFK |
| PyTorch | RePAD-PT | SALAD-PT |
| Deeplearning4j | RePAD-DL4J | SALAD-DL4J |

## 4.1 Real-world datasets

To evaluate the three RePAD combinations, two real-world time series were used. One is called ec2-cpu-utilization-825cc2 (CC2 for short), and the other is called rds-cpu-utilization-e47b3b (B3B for short). Both time series are provided by the Numenta Anomaly Benchmark (NAB) (Lavin and Ahmad, 2015). CC2 contains two point anomalies and one collective anomaly, whereas B3B contains one point anomaly and one collective anomaly. Note that a point anomaly is a single data point which is identified as anomalous with respect to the rest of the time series, whereas a collective anomaly is defined as a sequence of data points which together form an anomalous pattern (Schneider et al., 2021).

Since CC2 and B3B consist of only 4032 data points, they are unable to show the long-term performance of the three RePAD combinations. Hence, we created two long time series called CC2-10 and B3B-10 by individually duplicating CC2 and B3B ten times. Table 2 lists their details. Figures 1 and 2 illustrate all data points in CC2-10 and B3B-10, respectively. Each point anomaly is marked as a red circle, whereas each collective anomaly is marked as a red curve line.

Table 2: Two extended real-world time series used to evaluate RePAD-TFK, RePAD-PT, and RePAD-DL4J.

| Name | Number of data points | Time interval | Duration | Number of anomalies |
|---|---|---|---|---|
| CC2-10 | 40,320 | 5 | 140 days | 20 point and 10 collective anomalies |
| B3B-10 | 40,320 | 5 | 140 days | 10 point and 10 collective anomalies |

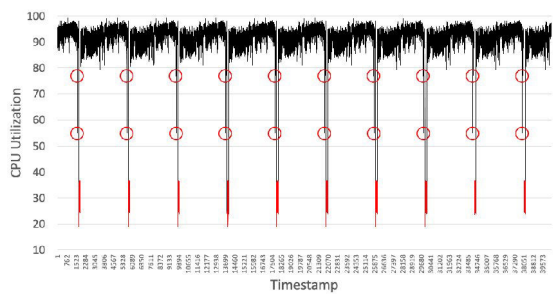

Figure 1: All data points on the CC2-10 time series. Each anomaly is marked in red.

On the other hand, to evaluate the three SALAD combinations, we selected another two real-world re-current time series. One is Taipei Mass Rapid Transit (TMRT for short) (Yeh et al., 2019), and the other is New York City Taxi demand (NYC for short) from the Numenta Anomaly Benchmark (Lavin and Ahmad, 2015). The former consists of 1260 data points, whereas the latter consists of 10320 data points. Table 3 summarizes the details of TMRT and NYC. They contain only collective anomalies.

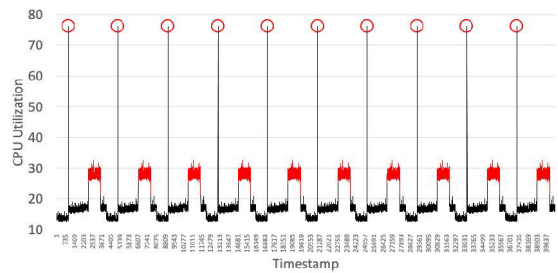

Figure 2: All data points on the B3B-10 time series. Each anomaly is marked in red.

Table 3: Two real-world time series used to evaluate SALAD-TFK, SALAD-PT, and SALAD-DL4J.

| Name | Number of data points | Interval | Duration | Number of anomalies |
|---|---|---|---|---|
| TMRT | 1,260 | 1 hour | 2016/02/01 00:00 to 2016/03/31 23:00 | 1 collective anomaly |
| NYC | 10,320 | 30 min | 2014/07/01 00:00 to 2015/01/31 23:30 | 5 collective anomalies |

## 4.2 Hyperparameters, parameters, and environment

To ensure a fair evaluation, the three RePAD combinations were configured with the same hyperparameters and parameters, as listed in Table 4, following the setting used by RePAD (Lee et al., 2020b). Recall that RePAD utilizes the Look-Back and Predict-Forward strategy to determine data size for online model training and data size for prediction. In this paper, we respectively set the Look-Back parameter and the Predict-Forward parameter to 3 and 1 based on the setting suggested by (Lee et al., 2021a). In other words, the LSTM models used by RePAD-TFK, RePAD-PT, and RePAD-DL4J will be always trained with three historical data points, and the trained models will be used to predict the next upcoming data point in the target time series.

In addition, RePAD-TFK, RePAD-PT, and RePAD-DL4J inherited the simple LSTM structure used by RePAD (Lee et al., 2020b), i.e., only one hidden layer and ten hidden units. Note that Early stopping (EarlyStopping, 2023) was not used to automatically determine the number of epochs since this technique is not officially supported by PyTorch.

For fairness, the number of epochs was set to 50 for the three RePAD combinations.

Table 4: The hyperparameter and parameter setting used by RePAD-TFK, RePAD-PT, and RePAD-DL4J.

| Hyperparameters/parameters | Value |
|---|---|
| The Look-Back parameter | 3 |
| The Predict-Forward parameter | 1 |
| The number of hidden layers | 1 |
| The number of hidden units | 10 |
| The number of epochs | 50 |
| Learning rate | 0.005 |
| Activation function | tanh |
| Random seed | 140 |

Table 5: The hyperparameter and parameter setting used by SALAD-TFK, SALAD-PT, and SALAD-DL4J.

| Hyperparameters/parameters | The conversion phase | The detection phase |
|---|---|---|
| The Look-Back parameter | 288 for NYC, 63 for TMRT | 3 |
| The Predict-Forward parameter | 1 | 1 |
| The number of hidden layers | 1 | 1 |
| The number of hidden units | 10 | 10 |
| The number of epochs | 100 | 50 |
| Learning rate | 0.001 | 0.001 |
| Activation function | tanh | tanh |
| Random seed | 140 | 140 |

Similarly, to make sure a fair evaluation, the three SALAD combinations were all configured with the same hyperparameter and parameter setting, as listed in Table 5. However, the setting is slightly different when it comes to the two used time series TMRT and NYC. Recall that SALAD consists of one conversion phase and one detection phase. The conversion phase requires more data points for model training than the detection phase does. Hence, the Look-Back parameter for the conversion phase of SALAD-TFK, SALAD-PT, and SALAD-DL4J were all set to 288 and 63 on NYC and TMRT, respectively. Due to the same reason, we configured 100 and 50 epochs for the conversion phase and the detection phase of the three SALAD combinations, respectively. On the other hand, the Look-Back parameter for the detection phase of the three SALAD combinations were all set to 3 no matter the used time series is TMRT or NYC. This is because the detection phase works exactly like RePAD, and three is the recommend value suggested by (Lee et al., 2021a) for the Look-Back parameter of RePAD.

The evaluations for all the six combinations were individually performed on the same laptop running MacOS 10.15.1 with 2.6 GHz 6-Core Intel Core i7 and 16GB DDR4 SDRAM. Note that we did not choose GPUs or high-performance computers to conduct the evaluation since it is interesting to know how TensorFlow-Keras, PyTorch, and Deeplearning4j impact RePAD and SALAD on a commodity computer.

## 5 EVALUATION RESULTS

In this section, we detail the evaluation results of the three RePAD combinations and the three SALAD combinations.

### 5.1 Three RePAD combinations

To measure the detection accuracy for each RePAD combination, we chose precision, recall, and F-score. Precision is the ratio between the true positives (TP) and all the positives, i.e., precision= TP/(TP+FP) where FP represents false positive. Recall is the measure of the correctly identified anomalies from all the actual anomalies, i.e., recall= TP/(TP+FN) where FN represents false negative. F-score is a well-known composite measure to evaluate the accuracy of a model, and it is defined as 2·(precision·recall)/(precision+recall). A higher value of F-score indicates better detection accuracy.

It is worth noting that we did not utilize the traditional pointwise approach to measure precision, recall, and F-score. Instead, we refer to the evaluation method used by (Lee et al., 2020a). More specifically, if a point anomaly occurring at time point $Z$ can be detected within a time period ranging from time point $Z-K$ to time point $Z+K$, this anomaly is considered correctly detected. On the other hand, for any collective anomaly, if it starts at time point $A$ and ends at time point $B$ ($B>A$), and it can be detected within a period between $A-K$ and $B$, we consider this anomaly correctly detected. In this paper, we set $K$ to 7 following the setting suggested by (Ren et al., 2019), i.e., $K$ is 7 if the measurement interval of a time series is a minute, and $K$ is 3 for a hourly time series.

In addition, we used three performance metrics to evaluate the efficiency of each RePAD combination. The first one is LSTM training ratio, which is the ratio between the number of data points that require a new LSTM model training and the total number of data points in the target time series. A lower ratio indicates less computation resources and quicker response time because LSTM model training takes some time. The second one is average detection time for each data point when LSTM model training is not required (ADT-NT for short). According to the design of RePAD, the LSTM model will not be replaced if it can accurately predict the next data point, which also means that the detection can be performed immediately without any delay. The last performance metric is average detection time when LSTM model training is required (ADT-T for short). When LSTM model training is required, the time to detect if a data point is anomalous consists of the time to train a new LSTM model, the time for this new model to re-predict the value of the data point, and the time to determine if the data point is anomalous. Apparently, ADT-T would be longer than ADT-NT due to LSTM model training.

Tables 6 to 9 show the performance of the three RePAD combinations on the CC2-10 time series. It is clear that RePAD-PT performs the best since it provides the highest detection accuracy, the least number of LSTM training, and the shortest ADT-T. The result shows that PyTorch seems to be a good choice for RePAD.

Although RePAD-TFK provides the second best detection accuracy, its ADT-NT and ADT-T were obviously the longest. It seems like TensorFlow-Keras is less efficient than PyTorch and Deeplearning4j.

On the other hand, we can see from Table 6 that RePAD-DL4J provides the lowest detection accuracy due to the lowest recall. Nevertheless, its ADT-NT is the shortest and its ADT-T is the second shortest with the smallest standard deviation. It seems that Deeplearning4j offers more stable execution performance than the other two libraries.

Table 6: The detection accuracy of the three RePAD combinations on the CC2-10 time series.

| Combination | Precision | Recall | F-score |
|---|---|---|---|
| RePAD-TFK | 0.957 | 0.9 | 0.928 |
| RePAD-PT | 0.954 | 0.934 | 0.944 |
| RePAD-DL4J | 0.964 | 0.7 | 0.811 |

Table 7: The LSTM training ratio of the three RePAD combinations on the CC2-10 time series.

| Combination | LSTM training ratio |
|---|---|
| RePAD-TFK | 0.0094 (379/40320) |
| RePAD-PT | 0.0089 (357/40320) |
| RePAD-DL4J | 0.0131 (528/40320) |

Table 8: The ADT-NT of the three RePAD combinations on the CC2-10 time series.

| Combination | ADT-NT (sec) | Std. Dev. (sec) |
|---|---|---|
| RePAD-TFK | 0.518 | 0.726 |
| RePAD-PT | 0.069 | 0.263 |
| RePAD-DL4J | 0.028 | 0.022 |

Tables 10 to 13 show the detection results of the three RePAD combinations on another time series B3B-10. Apparently, RePAD-TFK has the highest detection accuracy and the lowest LSTM training ratio. However, its ADT-NT and ADT-T are the longest. This result confirms that TenserFlow-Keras introduces more overhead to RePAD than the other two libraries do.

When RePAD was implemented in PyTorch, it has the second best detection accuracy, the second short-est ADT-NT, and the shortest ADT-T. In other words, PyTorch provides a very good balance between detection accuracy and response time. On the other hand, when RePAD-DL4J worked on B3B-10, its performance is similar to its performance on CC2-10 (i.e., the lowest detection accuracy but satisfactory execution performance).

Table 9: The ADT-T of the three RePAD combinations on the CC2-10 time series.

| Combination | ADT-T (sec) | Std. Dev. (sec) |
|---|---|---|
| RePAD-TFK | 1.913 | 1.409 |
| RePAD-PT | 0.100 | 0.318 |
| RePAD-DL4J | 0.375 | 0.030 |

Table 10: The detection accuracy of the three RePAD combinations on B3B-10.

| Combination | Precision | Recall | F-score |
|---|---|---|---|
| RePAD-TFK | 0.892 | 1 | 0.943 |
| RePAD-PT | 0.872 | 1 | 0.932 |
| RePAD-DL4J | 0.828 | 1 | 0.906 |

Table 11: The LSTM training ratio of the three RePAD combinations on B3B-10.

| Combination | LSTM training ratio |
|---|---|
| RePAD-TFK | 0.0026 (105/40320) |
| RePAD-PT | 0.0028 (112/40320) |
| RePAD-DL4J | 0.0042 (168/40320) |

## 5.2 Three SALAD combinations

To evaluate the detection accuracy of the three SALAD combinations, we also used precision, recall, and F-Score. Furthermore, we measured the average time for each SALAD combination to process each data point in their conversion phases and detection phases.

Figure 3 shows the detection results of the three SALAD combinations on the TMRT time series. Apparently, all of them can detect the collective anomaly without any false positive or false negative. Hence, the precision, recall, and F-score of the three combinations are all one as shown in Table 14.

Table 15 lists the time consumption of the three SALAD combinations on TMRT. It is clear that SALAD-PT has the shortest average conversion time and average detection time, whereas SALAD-TFK has the longest average conversion time and average detection time. It seems like PyTorch is also the best choice for SALAD so far.

Table 16 lists the detection results of the three SALAD combinations on the NYC time series. We can see that SALAD-DL4J has the best detection accuracy. Recall that the conversion phase of SALAD (Lee et al., 2021b) aims to convert a complex time series into a less complex AARE series by predicting the value for each future data point, measuring the difference between every pair of predicted and actual data points, and deriving the corresponding AARE values. As we can see from Figure 4 that most of the data points predicted by the conversion phase of SALAD-DL4J matched the real data points. Consequently, as shown in Figure 5, the detection phase of SALAD-DL4J was able to detect all the collective anomalies even though there are some false positives. However, the good performance of the conversion phase of SALAD-DL4J comes at the price of a long conversion time (see Table 17) due to required LSTM model training for many data points.

Table 12: The ADT-NT of the three RePAD combinations on the B3B-10 time series.

| Combination | ADT-NT (sec) | Std. Dev. (sec) |
|---|---|---|
| RePAD-TFK | 0.517 | 0.724 |
| RePAD-PT | 0.069 | 0.263 |
| RePAD-DL4J | 0.028 | 0.015 |

Table 13: The ADT-T of the three RePAD combinations on the B3B-10 time series.

| Combination | ADT-NT (sec) | Std. Dev. (sec) |
|---|---|---|
| RePAD-TFK | 1.989 | 1.436 |
| RePAD-PT | 0.105 | 0.325 |
| RePAD-DL4J | 0.388 | 0.039 |

On the other hand, when SALAD-TFK and SALAD-PT worked on NYC, they both had very poor detection accuracy (see Table 16). SALAD-TFK could detect only one collective anomaly, i.e., the snow storm. This is because the conversion phase of SALAD-TFK was unable to correctly predict data points (as shown in Figure 6). This bad performance consequently affected the detection phase of SALAD-TFK and disabled it to detect anomalies. We can see from Figure 7 that almost all AARE values are lower than the detection threshold.

If we look at Figure 7 more closely, we can see that the detection threshold was very high in the be-

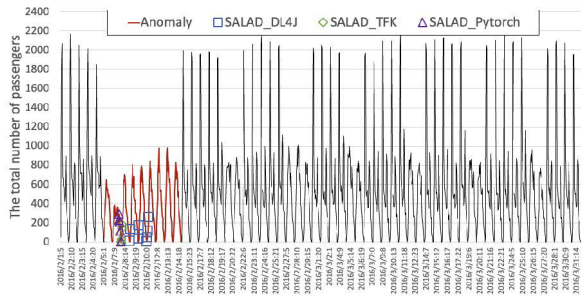

Figure 3: The detection results of the three SALAD combinations on the TMRT time series.

Table 14: The detection accuracy of the three SALAD combinations on the TMRT time series.

| Combination | Precision | Recall | F-score |
|---|---|---|---|
| SALAD-TFK | 1 | 1 | 1 |
| SALAD-PT | 1 | 1 | 1 |
| SALAD-DL4J | 1 | 1 | 1 |

Table 15: The time consumption of the three SALAD combinations on the TMRT time series.

| Combination | Average Conversion Time/Std. Dev.(sec) | Average Detection Time/Std. Dev. (sec) |
|---|---|---|
| SALAD-TFK | 0.949/1.017 | 0.472/0.703 |
| SALAD-PT | 0.023/0.163 | 0.008/0.087 |
| SALAD-DL4J | 0.162/0.399 | 0.011/0.027 |

Table 16: The detection accuracy of the three SALAD combinations on the NYC time series.

| Combination | Precision | Recall | F-score |
|---|---|---|---|
| SALAD-TFK | 0.447 | 0.2857 | 0.349 |
| SALAD-PT | 0.338 | 0.2857 | 0.310 |
| SALAD-DL4J | 0.709 | 1 | 0.830 |

Table 17: The time consumption of the three SALAD combinations on the NYC time series.

| Combination | Average Conversion Time/Std. Dev.(sec) | Average Detection Time/Std. Dev. (sec) |
|---|---|---|
| SALAD-TFK | 0.477/0.798 | 0.488/0.721 |
| SALAD-PT | 0.045/0.273 | 0.022/0.147 |
| SALAD-DL4J | 2.306/4.969 | 0.018/0.042 |

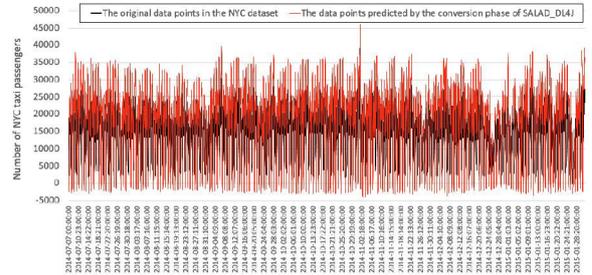

Figure 4: The original data points in the NYC time series versus the data points predicted by the conversion phase of SALAD-DL4J.

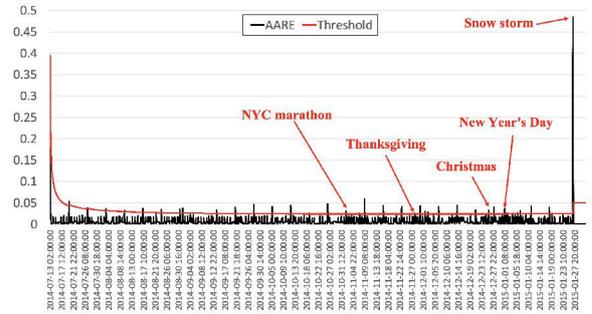

Figure 5: The AARE values generated by the detection phase of SALAD-DL4J versus the self-adaptive detection threshold of SALAD-DL4J on the NYC time series.

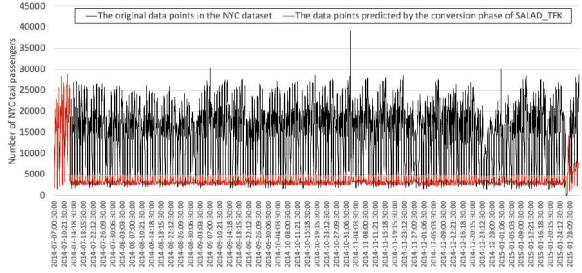

Figure 6: The original data points in the NYC time series versus the data points predicted by the conversion phase of SALAD-TFK.

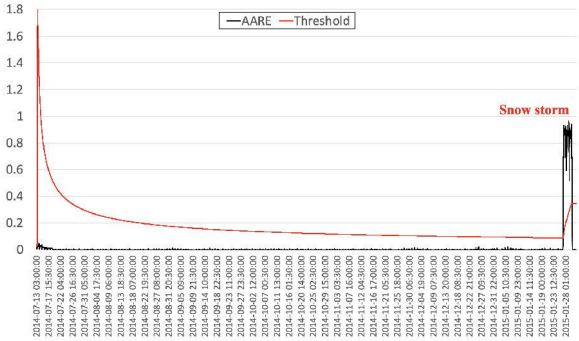

Figure 7: The AARE values generated by the detection phase of SALAD-TFK versus the self-adaptive detection threshold of SALAD-TFK on the NYC time series.

ginning due to the high AARE values, which makes SALAD felt that its current LSTM model did not need to be replaced. Even though the threshold dropped afterwards, it was still much higher than many subsequent AARE values. This is why most of the anomalies could not be detected. Since SALAD-TFK requires only a few model training, its average conversion time is much shorter than that of SALAD-DL4J (see Table 17).

The same situation happened to SALAD-PT when it worked on the NYC series. SALAD-PT has very poor detection accuracy even though its average conversion time and average detection time are the shortest.

## 6 CONCLUSIONS AND FUTURE WORK

In this paper, we investigated how DL libraries impact online adaptive lightweight time series anomaly detection by implementing two state-of-the-art anomaly detection approaches (RePAD and SALAD) in three well-known DL libraries (TensorFlow-Keras, PyTorch, and Deeplearning4j) and conducting a series of experiments to evaluate their detection performance and time consumption based on four open-source time series. The results indicate that DL libraries have a significant impact on RePAD and SALAD in terms of not only their detection accuracy but also their time consumption and response time.

According to the results, TensorFlow-Keras is not recommended for online adaptive lightweight time series anomaly detection because it might lead to unstable detection accuracy and more time consumption. When it was used to implement RePAD, RePAD had satisfactory detection accuracy. However, when it was used to implement SALAD, SALAD had unstable detection accuracy on one used time series. Besides, TensorFlow-Keras is less efficient than PyTorch and Deeplearning4j because it causes the longest response time for both RePAD and SALAD.

On the other hand, PyTorch is the most efficient library among the three DL libraries since it enables RePAD and SALAD to provide real-time processing and instant responses. It also enables RePAD to provide high detection accuracy. However, similar to TensorFlow-Keras, it causes unstable detection accuracy when it was used to implement SALAD and worked on the NYC time series.

Deeplearning4j is considered the most stable library among the three DL libraries because it not only enables RePAD and SALAD to provide satisfactory detection accuracy, but also enables RePAD and SALAD to have reasonable time consumption and response time.

We found that it is very important to carefully choose DL libraries for online adaptive lightweight time series anomaly detection because DL libraries might not show the true performance of an anomaly detection approach. What makes it even worse is that they might mislead developers or users in believing that one bad anomaly detection approach implemented in a good DL library is better than a good anomaly detection approach implemented in a bad DL library.

In our future work, we would like to release all the source code (i.e., RePAD and SALAD implemented in the three DL libraries) on a public software repository such as GitHub, GitLab, or Bitbucket.

## ACKNOWLEDGEMENT

The authors want to thank the anonymous reviewers for their reviews and suggestions for this paper.